\pdfoutput=1

\documentclass[11pt]{article}

\usepackage[preprint]{acl}

\usepackage{times}
\usepackage{booktabs, multirow, tabularx} 
\usepackage{latexsym}

\usepackage{listings}

\usepackage[T1]{fontenc}

\usepackage[utf8]{inputenc}

\usepackage{microtype}

\usepackage{inconsolata}

\usepackage{graphicx}

\usepackage{xcolor}   

\usepackage{amsmath}

\title{Addressing LLM Diversity by Infusing Random Concepts}

\author{Pulin Agrawal \\
  Pennsylvania State University \\
  \texttt{pja5407@psu.edu} \\\And
  Prasoon Goyal \\
  Amazon\\
  \texttt{prasog@amazon.com} \\}

\begin{document}
\maketitle
\begin{abstract}
Large language models (LLMs) are known to produce outputs with limited diversity. In this work, we study whether infusing random concepts in the prompts can improve the diversity of the generated outputs.
To benchmark the approach, we design a systematic evaluation protocol which involves prompting an LLM with questions of the form ``Name 10 Hollywood actors", and analyzing diversity measures of the resulting LLM outputs.
Our experiments on multiple LLMs show that prepending random words/sentences unrelated to the prompt result in greater diversity in the outputs of LLMs. 
We believe that this promising result and the evaluation protocol opens up interesting avenues for future work, such as how infusing randomness into LLMs could be applied to other domains. Further, the evaluation protocol could also inspire research into benchmarking LLM diversity more systematically.
\end{abstract}

\section{Introduction}

Over the last few years, large language models (LLMs) have led to unprecedented growth in artificial intelligence. These models are trained on massive amounts of data from which they learn facts, world models, and human opinions. These models can then make use of all the learned knowledge to solve novel tasks, and have been shown to be very effective at a wide variety of tasks such as chatbots, coding assistants, writing assistants, among others.

\begin{figure}
    \centering
    \includegraphics[width=0.85 \linewidth]{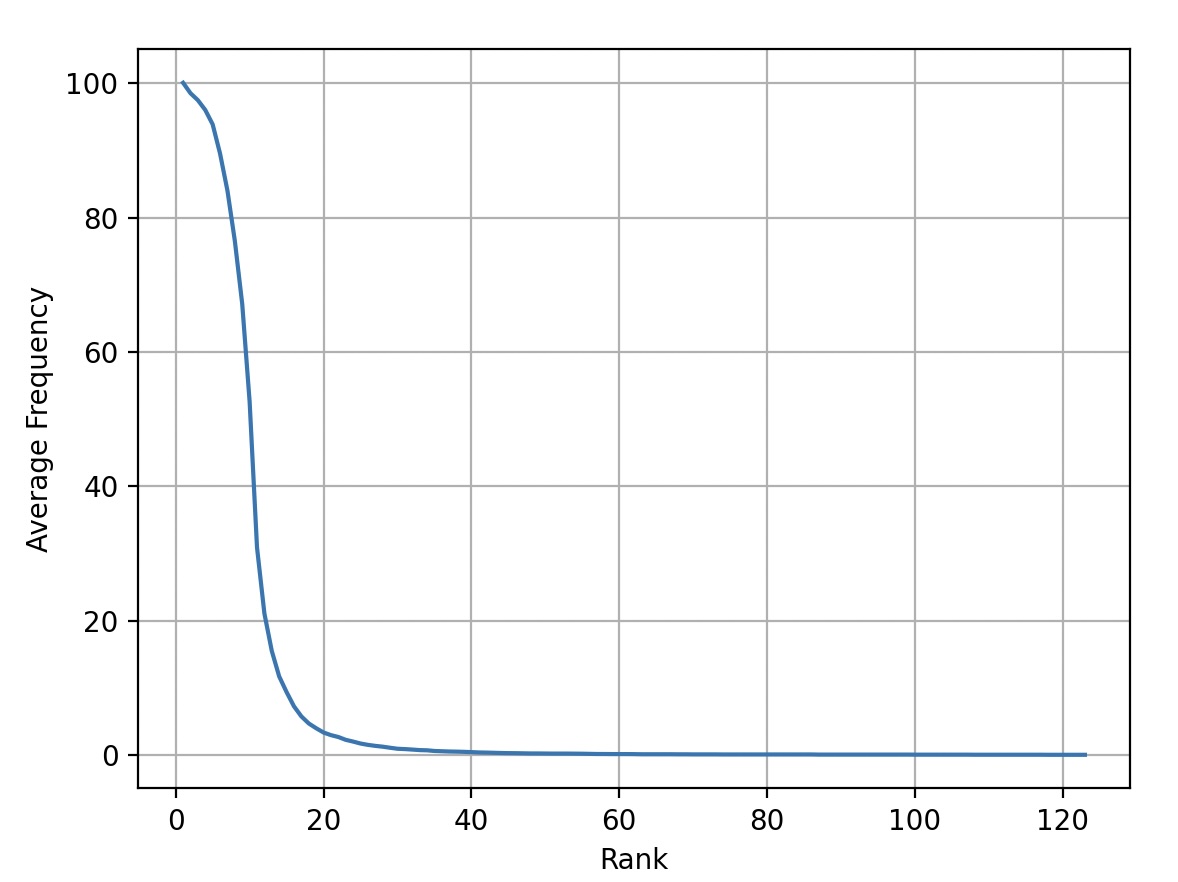}
    \caption{The long-tail problem of LLMs: Common responses have much higher frequency of generation by LLMs, while less common responses (on the long tail) are undersampled.}
    \label{fig:long-tail-distribution}
\end{figure}

However, the content generated by these models is often limited in terms of diversity.
For instance, consider a prompt like ``Name 10 Hollywood actors''. When an LLM is repeatedly asked this question, and the frequency distribution of the generated names is analyzed, it can be seen that the resulting distribution has low entropy, with a few names having very high frequency, some names having lower frequency, and most names having zero frequency. Figure \ref{fig:long-tail-distribution} demonstrates this effect, wherein the average frequency is much higher for higher rank items and a long tail of low frequency ranks has close to zero frequency.

We refer to this propensity of LLMs to oversample common responses, and undersample less common responses (i.e. on the ``long tail") as the \textbf{long-tail problem} of LLMs. It has also been referred to as the `mode collapse' problem in other works \citep{o2024attributing}. 
While this may be acceptable (or even desirable) in some applications, having higher output diversity is crucial in many applications.
Several prior works have noted this on different tasks, such as short story generation~\cite{doshi2024generative}, humor generation~\cite{jentzsch2023chatgpt}, and story completion~\cite{xu2024echoes}.
A number of approaches have been proposed to improve the diversity of LLM outputs. One line of work is based on modifying the sampling approach while decoding, such as top-k sampling~\cite{fan-etal-2018-hierarchical}, nucleus sampling~\cite{holtzman2020curious}, and logit suppression~\cite{chung-etal-2023-increasing}. Another class of approaches explores prompting the model in different ways \cite{lahoti2023improvingdiversitydemographicrepresentation,wang2025multilingual}.

In this work, we experiment whether LLMs can be made to generate more diverse outputs by injecting random concepts into the prompts.
We find that this simple idea is effective at generating more diverse outputs from LLMs.

While prior studies have qualitatively shown the effect of introducing randomness \citep{agrawal2024generative}, to the best of our knowledge, our work is the first to systematically study the impact of injecting randomness in the prompt on output diversity.

Our work is part of the body of work that studies prompt variation for increasing output diversity. However, most prior work requires generating paraphrases of given prompts, while our approach is very lightweight in comparison. Further, our approach is orthogonal to most other techniques for increasing diversity of LLM outputs, and can therefore be easily combined with them.

Our contributions in this paper are as follows:
\begin{itemize}
    \item We devise a light-weight prompting technique for increasing diversity of LLM outputs by infusing random concepts.
    \item We propose an evaluation methodology for systematically testing the diversity of responses generated by an LLM for list based questions. 
    \item Our experiments demonstrate effectiveness of this prompting technique in LLMs of various size and ranging from open-source to proprietary. 
\end{itemize}

\section{Infusing Random Concepts to Improve LLM Diversity}

We look at the problem of improving diversity in LLM outputs. In this work, we consider problems that require generating a list of K items from a much larger fuzzy set, and study the frequency distribution of the resulting LLM outputs. 

Specifically, consider a set of prompts $\{p_1, \ldots, p_N\}$, where each prompt is a question about generating a list of K items.
Given an LLM, each prompt $p_i$ is used to query the model $M$ times, generating lists $l_{j}=(x_j^1, \ldots, x_j^K)$, where $j=\{1, \ldots, M\}$.
We then create the concatenated list $L_{p_i} = (l_1, \ldots, l_M)$, and compute the number of unique elements and the entropy:
\[
\text{Count}({p_i}) = |set(L_{p_i})|
\]
\[
\text{Entropy}({p_i}) = H(L_{p_i}) = -\sum_{x \in set(L_{p_i})} p(x) \log p(x)
\]
where $p(x)$ is the empirical probability of element $x$ in $S$.


Intuitively, a higher count means that the LLM generates more unique responses for the given prompt. A higher entropy means that the frequency distribution of the responses has a heavier tail, suggesting less bias towards more common responses.

We experiment with adding random context to the prompt, and measure the change in count and entropy. Let $c_j$ be a random context, and $p_i$ be the prompt under consideration. We modify the prompt by prepending $c_j$ to $p_i$ for the $j^{th}$ query to the model:
\[
p_{ij} = c_j \circ p_i
\]
where $\circ$ denotes the string concatenation operator.
The modified prompts $p_{ij}$ are used to generate lists $l'_j$, which are concatenated to get $L'_{p_i} = (l'_1, \ldots, l'_M)$. Finally, we compute the modified statistics:
\[
\text{Count}'({p_i}) = |set(L'_{p_i})|
\]
\[
\text{Entropy}'({p_i}) = H(L'_{p_i})
\]


By measuring the difference between the statistics for the original responses $L'_{p_i}$ and the updated responses $L'_{p_i}$, we can make inferences about the effectiveness of our proposed approach.

\section{Experiments and Results}

\newcolumntype{C}{>{\centering\arraybackslash}p{1.4cm}}
\begin{table*}[ht]
\centering
\caption{Mean Entropy and Median Count  Comparison Across Models and Datasets}
\small
\begin{tabular}{llcccccc}
\toprule
\multirow{2}{*}{\textbf{Model}} & \multirow{2}{*}{\textbf{Dataset}} & \multicolumn{2}{c}{\textbf{Regular Prompt}} & \multicolumn{2}{c}{\textbf{With Random Word}} & \multicolumn{2}{c}{\textbf{With Random Sentence}} \\
\cmidrule(lr){3-4} \cmidrule(lr){5-6} \cmidrule(lr){7-8}
& & \textbf{Entropy} & \textbf{Count} & \textbf{Entropy} & \textbf{Count} & \textbf{Entropy} & \textbf{Count}\\
\midrule
Gemma3:4b & ordered          & 3.89  & 22 & 4.09 & 32 & 4.08 & 30 \\
          & unordered        & 3.85  & 18 & 4.16 & 35 & 4.16 & 31 \\
Nova Pro  & ordered          & 4.15  & 31 & 4.24 & 34 & 4.22 & 32 \\
          & unordered        & 4.13  & 31 & 4.25 & 33 & 4.21 & 33 \\
Claude 3.5 Sonnet  & ordered & 3.65  & 16 & 3.70 & 17 & 3.68 & 16 \\
          & unordered        & 3.65  & 16 & 3.69 & 18 & 3.66 & 17 \\
Mistral Large  & ordered     & 4.03  & 22 & 4.15 & 26 & 4.16 & 26 \\
          & unordered        & 3.93  & 21 & 4.08 & 26 & 4.14 & 28 \\
\bottomrule

\label{tab:summary-results}
\end{tabular}
\end{table*}

First, we generate a dataset of prompts using an LLM. We use an LLM to generate candidate prompts, which are manually filtered to remove duplicates and low-quality questions (see Appendix for examples). This gives us a set of $N=100$ prompts. Each prompt in the dataset requires generating a list of $K=10$ items from a fuzzy set. For each prompt, we generate responses from the model $M=100$ times.

We experiment with 2 types of settings. In the first one, we add a qualifier that can be used to loosely order the list of items in the fuzzy set. For instance, ``Name 10 popular Hollywood actors'', where the list of all Hollywood actors can be loosely ordered by their popularity.
In the second setting, we do not add this qualifier. For instance, ``Name 10 Hollywood actors.''
We refer to these two settings as ``ordered'' and ``unordered'' respectively.
Ordered queries contain an explicit bias, encouraging the model to select from a narrower and more stereotypical set of responses. In contrast, unordered queries are more open-ended and allow greater variation. By testing both ordered and unordered queries, we assess whether our method enhances diversity across both these settings, thereby testing its robustness in constrained settings.

For each setting, we generate outputs using an LLM, with and without prepending a random context. 
We experiment with 2 types of random contexts: (1) prepending a random word, and (2) prepending a random sentence.
We use the Python package {\tt wonderwords} to generate random words and sentences.


We refer to the baseline approach, without any random context added, as ``Regular Prompt'', while our approach where a random word and a random sentence is added to the prompt, as ``With Random Word'' and ``With Random Sentence" respectively.

We experiment with the following models of varying sizes, including open and closed models: (1) Gemma3 4b \cite{team2025gemma}, (2) Mistral-Large \cite{mistral2024large}, (3) Amazon Nova Pro \cite{intelligence2024amazon}, and (4) Claude 3.5 Sonnet \cite{anthropic2024claude35sonnet}. All models are used in structured output mode to produce JSON-like lists, with temperature set to 0.9 and other settings left at default. A summary of our results is shown in Table~\ref{tab:summary-results}. It can be seen that across all models and in both ``ordered'' and ``unordered'' settings, our approach results in a higher number of unique responses (i.e. count) as well as a greater entropy in LLM outputs. 

Having established the effectiveness of adding random context using various models, we next perform some additional analysis and experiments to better understand the impact of the approach. For these analysis and experiments, we focus on the Gemma3 4b model.

\begin{figure}

        \centering
    \includegraphics[width=0.95 \linewidth]{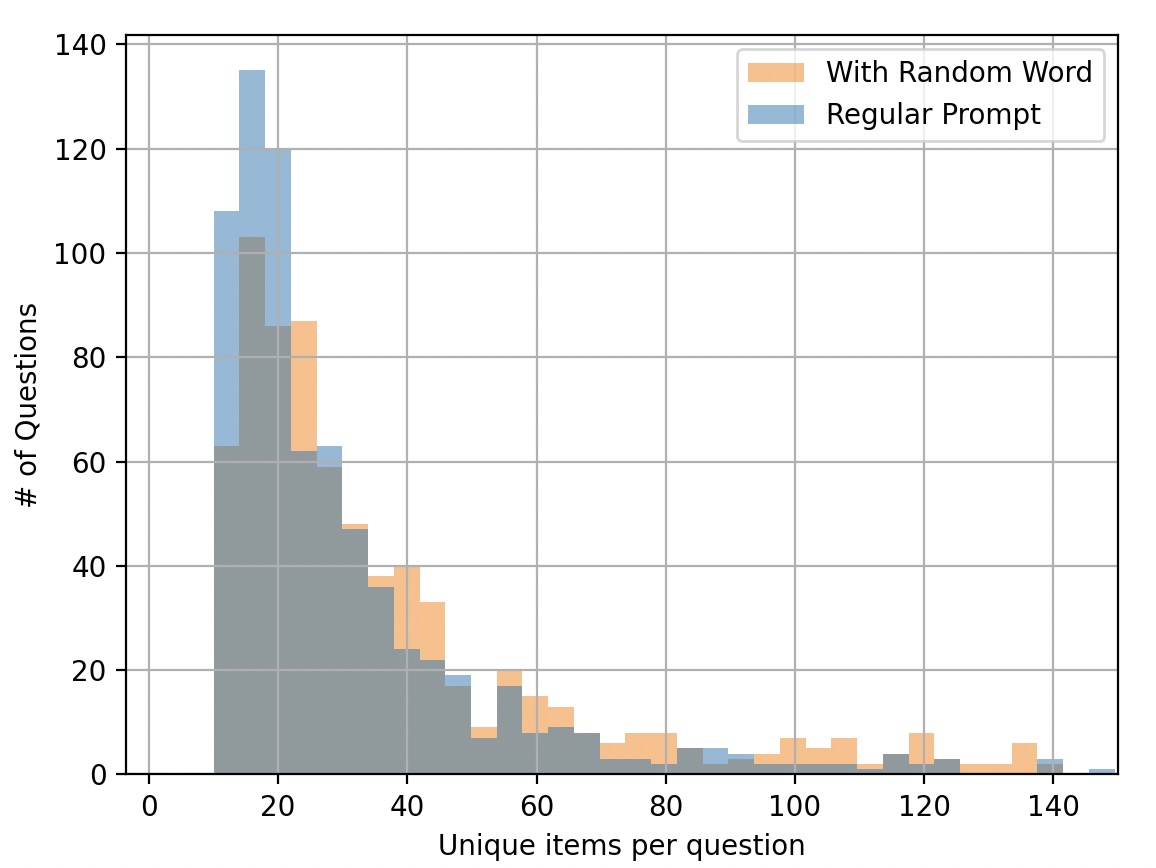}
    \caption{Histogram of unique items in the responses from all models with and without a random word context across the `ordered' and `unordered' settings.}
    \label{fig:unique-items-histogram}
\end{figure}

First, we visualize the histogram of unique items in the responses as shown in Figure \ref{fig:unique-items-histogram} for ``Regular Prompt" and ``With Random Word". We see an increase in the number of questions with more unique responses. 

\begin{figure}
    \centering
    \includegraphics[width=0.95 \linewidth]{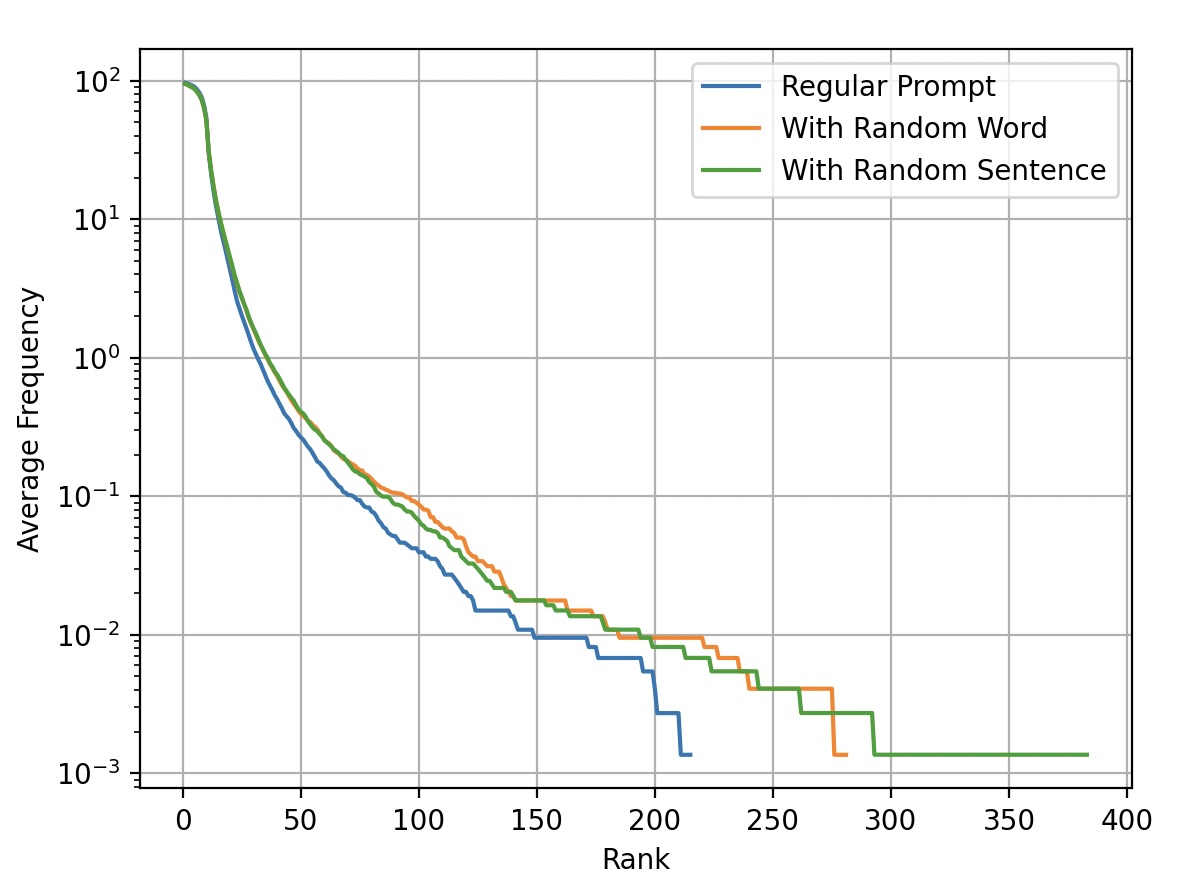}
    \caption{Average frequency vs. frequency ranks for all models across the `ordered' and `unordered' settings.}
    \label{fig:log-freq-rank}
\end{figure}

\begin{figure}
    \centering
    \includegraphics[width=\linewidth]{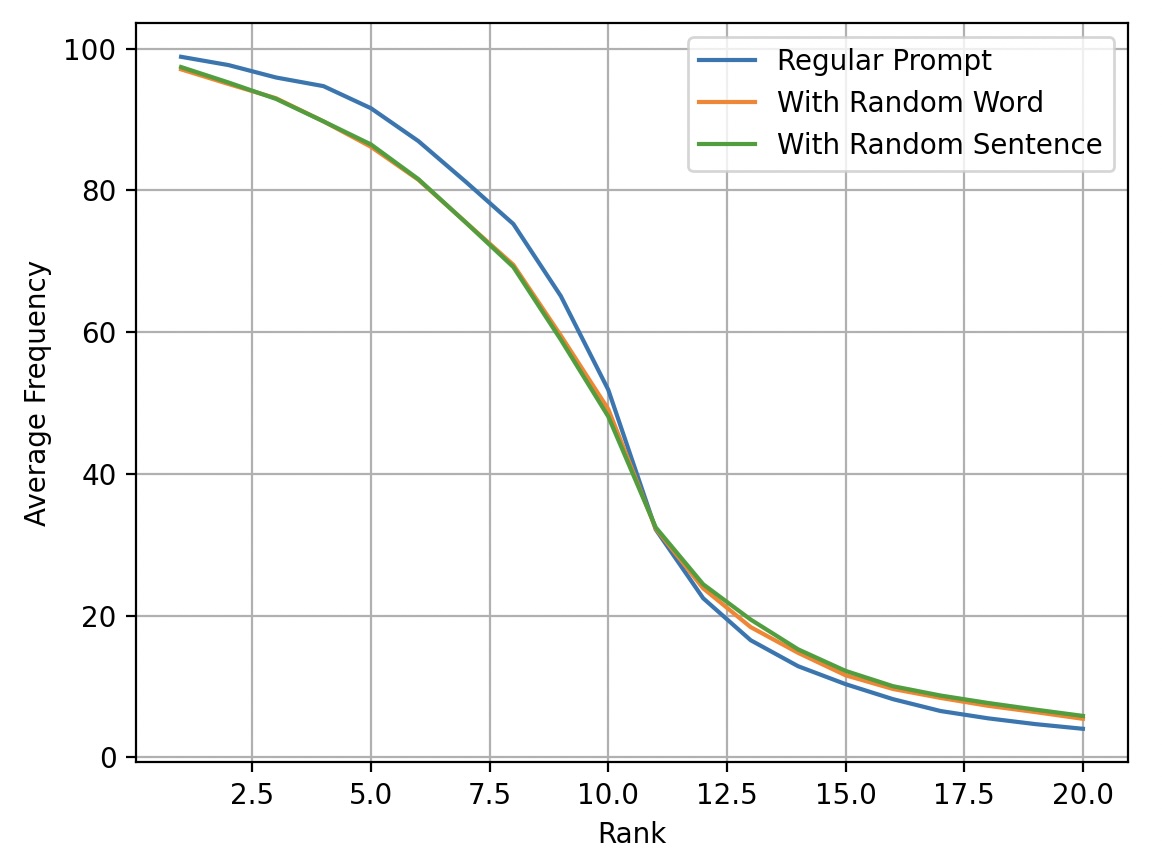}
    \caption{Average frequency vs. frequency ranks for Gemma3 4b model in unordered setting for higher ranks from Figure \ref{fig:log-freq-rank}.}
    \label{fig:high-rank-freq}
\vspace{-5mm}    
\end{figure}

Figure \ref{fig:log-freq-rank} shows average frequency for a given frequency rank.  Figure \ref{fig:high-rank-freq} shows high rank items from Figure \ref{fig:log-freq-rank}. For lower rank items, frequency is higher for random word prompt and random sentence prompt, whereas for higher rank items we can see a reverse trend (in Figure \ref{fig:high-rank-freq}). This indicates that our approach flattens the distribution to reduce the oversampling of common/frequent items. Since these results used temperature=0.9, it suggests this approach is orthogonal to temperature in promoting diversity.

We also compare the infusing multiple random words before the prompt. We added 1, 2, 5 and 10 words to understand the effect of more words on the entropy of the frequency distribution. The entropy results can be seen in Table \ref{tab:word-table}.


Further, we run paired t-tests on number of unique responses and entropy between the set of prompts ($N=100$) with and without random context (random word and random sentence). We find that in all settings for all models, the differences are statistically significant, at 5\% significance level.

We report additional results from some ablation experiments in Appendix~\ref{sec:ablations}.

\begin{table}[ht]
  \centering
  \caption{Entropy with varying number of random words in ordered setting for Gemma 3:4b}
  \small
  \begin{tabular}{l*{3}{cc}} 
    \toprule
     \textbf{\# of random words} & \textbf{Entropy} & \textbf{Count} \\
     \midrule
      1  &  4.09  & 32\\
      2  &  4.07  & 33\\
      5  &  4.15  & 34\\
      10 &  4.08  & 30\\
    \bottomrule
  \end{tabular}
  \label{tab:word-table}
\end{table}

\section{Discussion}

To address the limited diversity of LLM outputs, a problem we refer to as the ``long-tail'' problem, we proposed a lightweight approach that involves infusing a random concept (word/sentence) in the LLM prompt. Our experiments show that the proposed approach is effective at improving the LLM generations on questions that require generating lists of responses, resulting in statistically significant improvements both in the number of unique responses generated, as well as the entropy of the response distribution (Figure \ref{fig:unique-items-histogram} and \ref{fig:log-freq-rank}). 

Further, we did not observe a statistically significant increase in entropy when comparing prompts with different numbers of random words using paired t-tests or random word with random sentence settings. This may be indicative of saturation of effect of randomness.

We would like to conclude by discussing the broader theme of creativity, within which our proposed work fits.
Our work is inspired by psychological experiments~\cite{nemeth1985originality} which show that creativity of human subjects is enhanced when noise is introduced in the process.
This leads to the natural question: can we apply a similar idea to address the long-tail problem in LLMs?
Given that LLMs are trained on the next-word prediction task, it is not surprising that the conditional probability distributions learned by these models assign most of their probability density to a small set of high-frequency words occurring in the training set given a particular context.
By injecting randomness into their prompts, we are therefore changing these conditional probability distributions, leading to more diverse outputs.

A more thorough analysis from an explainability lens to better understand the mechanics of why injecting randomness into the prompt leads to more diverse outputs is left for future work.

Another follow-up question that arises is whether injecting random concepts could enable AI models to generate truly novel ideas, a long-standing goal of artificial intelligence.
As generative AI becomes more ubiquitous and gets integrated into everyday workflows of creative professionals like writers, artists, and scientists, this problem will become even more relevant.
As alluded to in the introduction, the long-tail problem of LLMs that we study in our work has also been observed in broader creative tasks, such as story and humor generation \cite{mirowski2024robot} and image generation \cite{dombrowski2025cvpr}. 
It would be interesting to study whether adding random concepts also improves LLM creativity in these tasks. However, post-hoc infusion of creativity in models may have its limits.
An important follow-up research question, therefore, is to explore further whether the current architectures and training techniques are sufficient to build models that can generate truly novel ideas, or if we need alternate architectures and/or training approaches.
Further, we also need to develop techniques to analyze more broadly the diversity of AI-generated content, which would be instrumental in answering the above research questions.

\section{Limitations}

While effective, our study has several limitations. 
First, even though the differences between ``Regular Prompt'' and ``With Random Word'' settings are statistically significant, the relative difference in number of unique responses and entropy of the response distribution are relatively small, especially in the ``ordered'' setting. As such, it is worth studying alternate approaches to infuse random concepts in the prompts to achieve greater relative change in LLM output distribution.
Second, the scope of our study is limited to list-based questions. While that makes the analysis tractable, it would be informative to study how well the proposed approach works in other settings. This would require defining alternate statistics to measure the diversity of LLM outputs, which we leave for future work.
Finally, in our experiments, we did not observe that adding random contexts to prompts lead to inferior outputs for the question in the prompt (e.g. the model generating irrelevant outputs), but it has been observed in prior work that there is generally a trade-off between diversity and accuracy. It would be interesting to study in the future how infusing random concepts could impact the accuracy/quality of the model outputs.

\clearpage

\bibliography{custom}

\clearpage
\appendix

\section{Appendix}
\label{sec:appendix}
\subsection{Ablation}
\label{sec:ablations}

We tested with random string (ex. `askfjiwubrc') instead of random word or random sentence. These distribution of the length of strings was similar to the random word scenarios in Table~\ref{tab:random-string-results}. For both the ordered and unordered setting we found a statistically significant result on introducing the random strings as compared to the control setting.

\begin{table}[ht]
\centering
\caption{Mean Entropy and Median Count Comparison for Gemma3}
\small
\setlength{\tabcolsep}{3pt} 
\renewcommand{\arraystretch}{1.5} 
\begin{tabular}{lcccc}
\toprule
 \multirow{2}{*}{\textbf{Dataset}} & \multicolumn{2}{c}{\textbf{Regular Prompt}} & \multicolumn{2}{c}{\textbf{Random String}} \\
\cmidrule(lr){2-3} \cmidrule(lr){4-5}
& \textbf{Entropy} & \textbf{Count} & \textbf{Entropy} & \textbf{Count} \\
\midrule
  ordered          & 3.89  & 22 & 4.09 & 32 \\
  unordered        & 3.85  & 18 & 4.16 & 35 \\

\bottomrule
\label{tab:random-string-results}
\end{tabular}

\end{table}

\subsection{Data}
This appendix sub-section provides examples of data used and responses generated by the models for experiments reported in this paper for context and clarity.
\\

\noindent Examples of questions in the \textbf{ordered set}.
\begin{itemize}
\item Name 10 best-known Shakespeare plays.
\item Name 10 most watched Olympic sports.
\item Name 10 largest African countries by area.
\item Name 10 major mountain ranges worldwide.
\item Name 10 most visited U.S. national parks.
\item Name 10 highest-grossing Marvel Cinematic Universe movies.
\item Name 10 Grammy-winning female vocalists.
\item Name 10 influential hip-hop producers.
\end{itemize}

\noindent Examples of questions in the \textbf{unordered set}.

\begin{itemize}
\item Name 10 Shakespeare plays.
\item Name 10 Olympic sports.
\item Name 10 African countries.
\item Name 10 mountain ranges worldwide.
\item Name 10 U.S. national parks.
\item Name 10 Marvel Cinematic Universe movies.
\item Name 10 female vocalists.
\item Name 10 hip-hop producers.
\end{itemize}

\noindent Example of \textbf{word randomness} and response.

\textbf{Question}: \\
curl clone complication giggle dangerous\\
Name 10 indigenous languages still spoken today.
\\

\textbf{Response}: \\
\lbrack'Navajo', 'Cherokee', 'Lakota', 'Ojibwe', 'Cree', 'Inuktitut', 'Zapotec', 'Quechua', 'Maya', 'Aymara'\rbrack
\\

\noindent Example of \textbf{sentence randomness} and response.

\textbf{Question}:\\
The sore enigma pauses eyestrain.\\
Name 10 fruits high in vitamin C.
\\

\textbf{Response}: \\
\lbrack'Acerola Cherry', 'Blackcurrant', 'Camu Camu', 'Citrus Fruits (Oranges, Grapefruit, Lemons, Limes)', 'Kiwi', 'Lychee', 'Mango', 'Papaya', 'Strawberries', 'Guava'\rbrack
\\

\subsection{Artifact Terms of Use}
To the best of our knowledge our use of language models adheres to the terms of use policy of each of the models intended use. All the code and data was generated by the authors with some help from AI.

\subsection{Model Size and Budget}
Model sizes are mentioned in Section 3 and total cost of running the experiments was <\$1000.

\subsection{AI Use}
ChatGPT, GitHub Copilot and other language models were used to generate and fix some code created for running these experiments and also to rephrase some text for writing of this paper.

\end{document}